%% file: main.tex
\documentclass[runningheads]{llncs}
\usepackage[T1]{fontenc}

\usepackage{graphicx}
\usepackage[most]{tcolorbox}  
\usepackage{placeins}
\usepackage{adjustbox}
\usepackage{tabularx}
\usepackage{xcolor}
\usepackage{float}
\usepackage{multirow}
\usepackage{makecell}
\usepackage{tcolorbox}
\tcbuselibrary{skins, breakable, listings}
\usepackage{caption}

\definecolor{pink}{rgb}{0.95, 0.8, 0.96}
\definecolor{pinkd}{rgb}{0.95, 0.4, 0.96}
\usepackage[textsize=tiny,backgroundcolor=pink,linecolor=pinkd]{todonotes}
\usepackage{hyperref}

\begin{document}

\title{OntoDoc: Automatically Generating Ontology Resource Descriptions}

\titlerunning{OntoDoc: Automatically Generating Ontology Resource Descriptions}

\author{Juan Guajardo Gutierrez, Mike Tran, Nasim Shirvani Mahdavi, Chengkai Li}
\institute{University of Texas at Arlington, Arlington TX 76013, US}
\authorrunning{J. Guajardo Gutierrez et al.}

\maketitle              

\input{sec-abstract}

\input{sec-introduction}
\input{sec-related}
\input{sec-method}
\input{sec-implement}
\input{sec-experiments}
\input{sec-results}
\input{sec-conclusion}

\paragraph*{Supplemental Material Statement:} Source code is available at
\url{https://anonymous.4open.science/r/ontodoc/}. Prompt templates and
evaluation scoring rubrics are located in the annex. System demo videos, the
raw data used to generate Tables~\ref{tab:model_selection},
\ref{tab:kg_assumptions}, \ref{tab:ontodoc-naive}, \ref{tab:auto-score-per-kg},
and~\ref{tab:human-eval}, and the ontology files and SPARQL endpoints used as
input to OntoDoc are attached with the submission in EasyChair and, if accepted,
will be made publicly available via a GitHub repository.
\input{sec-genai}

\bibliographystyle{splncs04}
\bibliography{refs}

\clearpage
\onecolumn
\input{appendix.tex}

\end{document}

%% file: sec-abstract.tex
\begin{abstract}

Knowledge graphs and ontologies form the backbone of structured, machine-readable knowledge representation, yet ontology terms often lack natural language descriptions, limiting their accessibility and reuse. Existing ontology verbalization approaches tend to rely on rule-based or template-driven techniques that operate on strict, formal schema axioms, resulting in rigid verbalizations that fail to convey richer contextual semantics. We present OntoDoc, a multi-agent framework for automatically generating natural language descriptions for any Internationalized Resource Identifier (IRI) within a SPARQL endpoint or RDF file. OntoDoc operates through a four-stage pipeline: (1) a Database Context agent that adaptively explores the knowledge graph via SPARQL to collect semantically informative triples about the target IRI; (2) a Web Context agent that retrieves and re-ranks encyclopedic background knowledge from Wikipedia and the open web;
(3) an Assumption Generation agent that synthesizes both contexts
into a structured, evidence-grounded list of intermediate assumptions; and (4) a Description Generation agent that produces the final
natural language description from the grounded assumptions. We evaluate OntoDoc on three knowledge graphs spanning biomedical, agricultural, and cross-domain settings, using both automatic scoring via Large Language Models (LLMs) and human evaluation with 11 annotators.
OntoDoc consistently matches or exceeds a naive LLM baseline and a template-based ontology verbalizer in accuracy. Furthermore, it vastly outperforms both baselines in the completeness of the generated descriptions, delivering an average 12\% improvement in completeness over the naive baseline across all domains, and improving completeness scores by as much as 2.8 points on a 5-point scale relative to the ontology verbalizer.

\keywords{Ontology verbalization  \and Knowledge graph documentation \and Large language models \and LLM agentic framework.}
\end{abstract}

%% file: sec-introduction.tex
\section{Introduction}

Knowledge graphs have become a foundational framework  for representing structured information in scientific, business and governmental applications~\cite{10.5555/2786071.2786081,10.1093/database/baab069}, as evidenced by initiatives such as the Proto-OKN program of the U.S. National Science Foundation~\cite{nsf_proto_okn}. Yet the practical utility of such semantic richness is contingent upon users' ability to comprehend and reuse the data. The semantics of a knowledge graph are largely defined by its ontology, which specifies the classes, properties, and relationships used to model the domain. In a typical knowledge graph, the ontology provides the intensional vocabulary, while instance data populates that vocabulary. Because the ontology defines the meaning of every term that appears in the data, it is the primary bottleneck for data comprehension. Users therefore tend to rely on ontology annotations such as \texttt{rdfs:comment}~\cite{w3c-rdfs-comment} to understand what each term means and how it should be used.

However, in practice, a large fraction of ontology terms carry no human-readable documentation, a problem visible across many publicly available \sloppy SPARQL endpoints, where \texttt{rdfs:comment} or equivalent annotations are frequently missing. This leaves users and downstream applications without any guidance about the intended meaning or usage of key graph concepts. The standard remedy, where a knowledge graph maintainer or domain expert inspects the graph, examines representative data, and writes a short description for each term, is not scalable. Large knowledge graphs can contain hundreds of distinct classes and properties~\cite{wikidata,dbpedia}, and the expertise required to describe them accurately is often concentrated in a small group of contributors, making systematic manual annotation infeasible. As a result, many knowledge graphs are deployed with incomplete documentation, hindering discovery, integration, and reuse.

Automatically generating ontology descriptions is therefore an important and practical problem. Existing approaches, however, remain limited. Traditional ontology verbalization systems primarily rely on schema-level axioms and hand-crafted linguistic templates to produce descriptions from formal ontology constructs~\cite{kaljurand2007verbalizing,stevens2011automating,third2011owl,cregan2007sydney,power2010complexity,androutsopoulos2013generating}. While effective for highly structured ontologies with rich logical axiomatization, these methods often generate rigid or overly generic descriptions and struggle when applied to knowledge graphs whose semantics are expressed implicitly through instance data rather than explicit ontology axioms. More recent approaches integrate Large Language Models (LLMs) to improve textual fluency~\cite{Zaitoun_Sagi_Peleg_2024}; however, they largely act as semantic translators of existing structure and remain strictly bounded by explicit schema axioms. Conversely, a natural first instinct is to directly prompt an LLM with raw graph triples and ask it to explain the target term. In practice, however, such naive prompting often yields incomplete descriptions because the semantics of ontology terms are rarely recoverable from a single local graph neighborhood.

To address these challenges, we present OntoDoc, an agentic multi-stage system for automatic ontology description generation. OntoDoc combines adaptive SPARQL-based graph exploration with external web retrieval and an intermediate reasoning stage that synthesizes evidence into structured assumptions before description generation. Rather than relying solely on schema axioms or fixed graph queries, OntoDoc actively explores the knowledge graph to gather semantically informative evidence, supplements sparse graph context with external encyclopedic knowledge, and grounds the final description in explicitly reasoned assumptions derived from both sources.

We evaluate OntoDoc on three diverse knowledge graphs spanning biomedical, agricultural, and open-domain data. Our evaluation includes both LLM-based and human assessment using four quality dimensions: accuracy, clarity, completeness, and cruciality. Experimental results show that OntoDoc consistently produces more complete and robust descriptions, improving completeness scores by up to 0.6 points on a 5-point Likert scale over the closest baseline while maintaining competitive accuracy and clarity across all domains. 

The main contributions of this work are as follows:\vspace{-6pt}

\begin{itemize}
    \item We formalize the ontology description generation task for heterogeneous knowledge graphs and define a multi-dimensional evaluation framework grounded in \texttt{rdfs:comment} quality criteria.
    \item We propose OntoDoc, an agentic multi-stage pipeline that combines adaptive graph exploration, external web retrieval, structured assumption generation, and LLM-based description synthesis.
    \item We introduce an intermediate assumption generation stage that improves factual grounding and reduces noise before final description generation.
    \item We evaluate OntoDoc across three real-world knowledge graphs and show improvements over existing baselines.
    \item All code used to produce this work is available in our anonymous GitHub repository at \href{https://anonymous.4open.science/r/ontodoc/}{https://anonymous.4open.science/r/ontodoc/}.
\end{itemize}

%% file: sec-related.tex
\section{Related Work}

\textbf{Ontology verbalization.}\hspace{2mm} 
A substantial body of research has investigated OWL~\cite{owl-spec} ontology verbalization~\cite{kaljurand2007verbalizing,stevens2011automating,third2011owl,cregan2007sydney,power2010complexity,androutsopoulos2013generating}. These approaches aim to transform formal OWL axioms into human-readable text, typically focusing on ontology classes or individuals. Most existing methods assume the presence of a well-defined schema graph and operate exclusively on the logical axioms expressed in OWL, rather than considering instance data or external context. Furthermore, they predominantly rely on pre-LLM techniques, including rule-based systems, logical axiom patterning, and Controlled Natural Language (CNL) ~\cite{kuhn-2014-survey} generation. While structurally precise, such approaches tend to produce rigid descriptions that closely mirror the formal, logical structure of the axioms (e.g., ``Class $A$ is a type of Class $B$ and is distinct from Class $C$)'', and fail to convey the intuitive real-world meaning of a concept. In contrast, OntoDoc generates descriptions grounded in both schema and instance data, and enriched with external knowledge, to produce more fluent and contextually rich explanations.

Zaitoun et al.~\cite{Zaitoun_Sagi_Peleg_2024} introduced an LLM-assisted verbalizer that converts OWL statements from existing ontologies into natural language, with the goal of generating text-to-OWL sentence pairs for fine-tuning ontology learning models. Their system uses a three-stage pipeline: initialization, CNL generation from OWL axioms, and an LLM-based paraphrasing step to improve fluency. The approach has the fundamental limitation of being strictly bounded by the explicit axioms present in a local ontology file. The system functions as a semantic translator of existing schema structure and cannot incorporate instance-level evidence from a live knowledge graph or enrich descriptions with external world knowledge. In contrast, OntoDoc is not limited to OWL ontologies nor constrained by predefined schema vocabularies. Rather than solely verbalizing logical axioms, OntoDoc integrates structural signals from the knowledge graph with contextual reasoning capabilities of pretrained LLMs, optionally incorporating relevant web-based information. This generates more informative and semantically enriched descriptions that extend beyond template-based verbalization.

\textbf{LLM-assisted ontology engineering.}\hspace{2mm}
Researchers have explored the use of LLMs to support ontology creation and engineering. For example, Lippolis et al.~\cite{lippolis2025ontology} investigated the generation of OWL ontology drafts from user stories and competency questions. Funk et al.~\cite{funk2023towards} proposed a method for constructing concept hierarchies across domains through LLM prompting. Similarly, Val et al.~\cite{val2025ontogenix} introduced OntoGenix, an LLM-powered pipeline for automating ontology development from complex datasets. While these approaches demonstrate the potential of LLMs in ontology construction, their primary objective is to generate or extend ontology structure. In contrast, OntoDoc focuses on enriching existing knowledge graphs with high-quality natural language descriptions, rather than synthesizing ontology structures from scratch.

\textbf{Graph-to-text generation.}\hspace{2mm}
A related but distinct line of research examines the use of LLMs and neural models to generate natural language descriptions from knowledge graphs~\cite{shi2023hallucination,zhang2023enhancing,du2024bi}. These graph-to-text approaches typically aim to transform subgraphs that are often centered on specific entities into coherent textual summaries. Crucially, these systems focus on describing data instances (e.g., summarizing facts about a specific person or event), whereas OntoDoc is designed to describe ontology terms (classes and properties), abstracting over their instances to define the term's intended meaning.

%% file: sec-method.tex
\section{Problem Formulation, Challenges and Methodology}
\label{sec:methodology}

{\flushleft \textbf{Problem Definition}.}\hspace{2mm}
Let $G = \{(s, p, o)\}$ be a knowledge graph represented as a set of RDF triples, where $s$ is the subject, $p$ is the predicate, and $o$ is the object (either an IRI or RDF literal; blank nodes are omitted). The graph is accessible via a SPARQL endpoint or a local RDF file. An IRI $u$ exists in $G$ if it appears as a subject, predicate, or object in at least one triple.

An IRI $u$ may play one of three semantic roles in $G$: class (explicitly declared as \texttt{owl:Class} or appearing as the object of \texttt{rdf:type} assertions), property (explicitly appearing as a predicate), or instance (typed to a class). OntoDoc makes no assumption about which role $u$ occupies and handles all cases uniformly.

Given a knowledge graph $G$ and a target IRI $u \in G$, the ontology description generation task is to produce a natural language string $d$ such that $d$ accurately and concisely describes the intended semantics of $u$ within the domain modeled by $G$, suitable for use in an \texttt{rdfs:comment}~\cite{w3c-rdfs-comment} or similar annotation.

{\flushleft \textbf{Overview of challenges}.}\hspace{2mm} 
Two properties of knowledge graphs make this task non-trivial: the semantics of an IRI are often only partially recoverable from its immediate graph neighborhood (understanding a class may require inspecting its instances and their attributes), and the wide variation in schema richness, instance density, and naming conventions means that no fixed retrieval strategy is reliably informative across domains.

{\flushleft \textbf{System Overview}.}\hspace{2mm} 
OntoDoc is designed to address these challenges through a central intuition: \textit{A good description of any ontology term should be grounded in the actual data it represents, while also offering intuitive guidance for both domain experts and first-time users.} 
This aligns with the W3C definition of \texttt{rdfs:comment}: ``A textual comment helps clarify the meaning of RDF classes and properties \ldots\ to indicate the intended meaning of the classes and properties described in an RDF vocabulary''~\cite{w3c-rdfs-comment}.

\begin{figure}[t]
    \centering
    \includegraphics[width=\columnwidth]{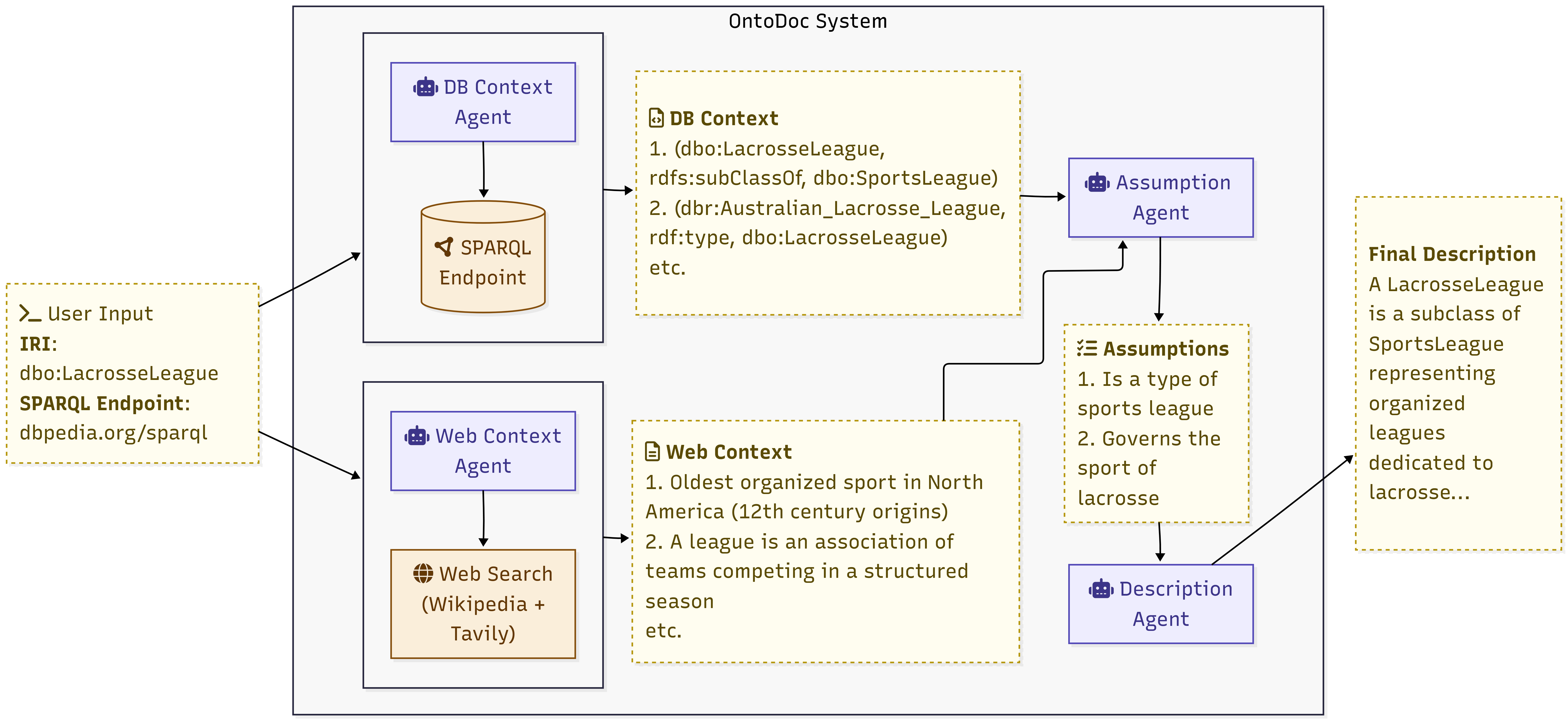}
    \caption{OntoDoc System Overview}
    \label{fig:ontodoc-system}
\end{figure}

As illustrated in Figure~\ref{fig:ontodoc-system}, OntoDoc comprises four components that execute in a fixed sequential pipeline: (1)~\textbf{database context extraction}, (2)~\textbf{web context retrieval}, (3)~\textbf{assumption generation}, and (4)~\textbf{description generation}. To reduce runtime, the database and web context stages run in parallel, and their outputs are then merged and passed downstream.

Each component is implemented as an autonomous LLM agent. Agents communicate exclusively through a shared session state in the form of a key-value store that persists across the pipeline run. No agent calls another directly, instead, each agent reads its required inputs from session state and writes its output back to a designated key (e.g., \texttt{database\_context}, \texttt{web\_context}, \texttt{assumptions}, \texttt{description}). This decoupled design allows components to be swapped, ablated, or replaced independently without modifying the rest of the pipeline. Full implementation details are provided in Section~\ref{sec:implement}.

%% file: sec-implement.tex
\section{Design and Implementation Details} \label{sec:implement}

\subsection{Database Context Agent} \label{sec:db_context}

The purpose of the database context agent is to gather the most semantically informative evidence available in the knowledge graph for the target IRI. Because a single predefined SPARQL query cannot capture the diverse shapes of classes, properties, and instances across varying ontologies, we implement the agent as an agentic SPARQL explorer that automatically selects and sequences queries within a configurable budget (default: $50$ triples, $10$ tool calls). The agent is equipped with five tools, summarized in Table~\ref{tab:tools}.

\setlength{\tabcolsep}{5pt}
\begin{table}[t]
\centering
\small
\caption{Agent toolset for ontology exploration and SPARQL querying.}
\resizebox{\columnwidth}{!}{%
\begin{tabular}{p{3.2cm} p{10cm}}
\hline
\textbf{Tool} & \textbf{Description} \\
\hline
\texttt{get\_iri\_information} 
& Retrieves rdf:type, rdfs:label, rdfs:comment, dcterms:description, and skos:definition; also checks whether the IRI is used as a predicate. \\
\texttt{get\_outgoing\_triples} 
& Returns triples where the IRI is subject ($\langle iri \rangle\ ?p\ ?o$), max 200. \\
\texttt{get\_incoming\_triples} 
& Returns triples where the IRI is object ($?s\ ?p\ \langle iri \rangle$), max 200. \\
\texttt{get\_predicate\_usage} 
& For property IRIs, returns sample triples ($?s\ \langle iri \rangle\ ?o$) and distinct rdf:type values of subjects and objects. \\
\texttt{execute\_sparql} 
& Executes SPARQL SELECT/ASK queries with filtering, grouping, aggregation, and arbitrary patterns. \\
\hline
\end{tabular}}
\label{tab:tools}
\end{table}

The agent follows a four-phase protocol embedded in its system prompt (detailed in Figure \ref{fig:db-context-prompt} in the Annex):

\textbf{Phase~1: Orient.}\hspace{2mm} 
The agent calls \texttt{get\_iri\_information}
on the target IRI to determine its type: \emph{class}
(\texttt{owl:Class}/\texttt{rdfs:Class} or many incoming
\texttt{rdf:type} triples), \emph{property} (appears as a predicate or
declared as \sloppy\texttt{owl:ObjectProperty}/\texttt{owl:DatatypeProperty}),
\emph{instance} (typed to a domain class), or \emph{unknown}.

\textbf{Phase~2: Initial scan.}\hspace{2mm}
The agent issues a sequence of queries depending on the type of the IRI. For a \emph{class}, it retrieves outgoing and incoming triples and executes a \texttt{COUNT} of instances; if the count is small ($<10$), all instances are fetched explicitly. Critically, the agent then
immediately inspects 2--3 individual instance IRIs via
\texttt{get\_outgoing\_triples} to observe the actual attributes and
relationships those instances carry. For a \emph{property}, it calls
\texttt{get\_predicate\_usage} to see what subject/object pairs it
connects, then inspects a representative subject and object to understand
the entities being linked. For an \emph{instance}, it retrieves outgoing
attributes and incoming references.

\textbf{Phase~3: Targeted follow-up.}\hspace{2mm}
The agent examines all results
accumulated so far and selects semantically significant IRIs (connected
classes, frequently occurring properties, shared object values) for
further exploration. For each candidate, it calls \texttt{get\_iri\_information} and, for classes, inspects some of their instances. Arbitrary SPARQL queries are used for counts of distinct values and predicate enumerations.

\textbf{Phase~4: Summarize.}\hspace{2mm}
The agent compiles a numbered, deduplicated list of triples in the canonical format:
$i.\ \langle subject \rangle\ \langle predicate \rangle\ \langle object \rangle$, where $i$ denotes the triple index in the list.
 
Deduplication rules prevent the output from being overwhelmed by
repetitive \texttt{rdf:type} assertions; at most two or three
representative triples are kept for any structural pattern. Pure
ontology axioms (\texttt{rdf:type owl:Class},
\texttt{owl:imports}, etc.) are omitted. The resulting list is written to the shared pipeline state under the key \texttt{database\_context}.

\subsection{Web Context} \label{sec:web_context}
Ontology terms frequently encode domain concepts whose meaning is not fully recoverable from the graph alone. The web context agent enriches the pipeline with external encyclopedic knowledge sourced from Wikipedia and the open web. The agent first extracts the human-readable label from the target IRI by splitting on the last \texttt{/} or \texttt{\#} character (e.g., 
\texttt{http://dbpedia.org/ontology/LacrosseLeague} $\to$ \texttt{LacrosseLeague}).  
The agent then applies a two-stage filter. First, labels that are trivially unambiguous English words (name, type, id, date, etc.) are rejected immediately without any retrieval call. This rejection is handled via a prompt-embedded heuristic where the agent is given an explicit enumeration of such stopword-like terms and instructed to match against it before proceeding. Second, for all other labels, the agent decomposes the label into searchable keywords and selects from the following two retrieval tools. The agent may invoke one or both, and their outputs are concatenated into a single context string stored in the shared pipeline under \texttt{web\_context}.

\textbf{\texttt{fetch\_wikipedia\_context}} is used for general, academic, or common terms likely to have a dedicated Wikipedia article. Results are retrieved via the Wikipedia Search API and then re-ranked by a text embedding model (\texttt{Cohere  Rerank 3.5} \cite{cohere_rerank_3_5_2024}) via API, keeping at most $k$ results (default $k=3$) above a configurable relevance threshold (default: $0.75$). This prevents low-quality, tangentially related articles from polluting the context.

\textbf{\texttt{fetch\_tavily\_context}} is used for highly technical, niche, or domain-specific terms unlikely to have a Wikipedia entry. It queries the Tavily web search API, which indexes technical documentation, scientific databases, and other non-encyclopedic sources. The choice between tools is determined entirely by the agent's own judgment about the label's nature, guided by examples provided in the system prompt (see Figure \ref{fig:web-context-prompt}).

\subsection{Assumption Generation} \label{assumps_generation}

The assumption generation agent receives the database context, web context, and target IRI, and is instructed to produce a JSON object of the following form: \texttt{\{ "items": [\{reasoning, assumptions\}, ...] \}}. Table~\ref{tab:lacrosse-assumptions} shows an excerpt of the assumptions and their corresponding reasoning as produced by the agent for the DBpedia class \texttt{LacrosseLeague}. The extracted assumptions are stored in the shared pipeline under the key \texttt{assumptions} and passed to the description generation agent.

\setlength{\tabcolsep}{5pt}
\begin{table}[t]
\centering
\small
\caption{Extracted assumptions and reasonings for \texttt{dbo:LacrosseLeague}}
\resizebox{\columnwidth}{!}{%
\begin{tabular}{p{7cm} p{7cm}}
\hline
\textbf{Assumption} & \textbf{Reasoning} \\
\hline
LacrosseLeague is a class representing organized sports leagues specifically for the sport of lacrosse.
& The IRI is typed as an OWL Class, specifically a subclass of \texttt{SportsLeague}, indicating that it represents a type of sports league focused on lacrosse. This is directly stated in triples 1 and 3. \\
\hline
LacrosseLeague is an internationally recognized concept describing leagues, associations, or organizations where lacrosse teams compete.
& Multiple labels in different languages such as `lacrosse league', `lacrosse bond', `Lacrosse-Liga' and others indicate the concept is universally recognized as a lacrosse sporting league entity. This is supported by triples 3--9. \\
\hline
\multicolumn{2}{c}{\ldots} \\
\hline
\end{tabular}}
\label{tab:lacrosse-assumptions}
\end{table}

The core idea is to force the model to reason explicitly before it writes. This is directly inspired by chain-of-thought prompting ~\cite{wei2022chain}, which demonstrates that eliciting intermediate reasoning steps substantially improves factual accuracy in downstream generation. By committing to a structured list of explicitly justified assumptions, the model is less likely to conflate evidence from different sources or hallucinate attributes not present in the graph. The reasoning field also makes the final description traceable to specific triples from the database context. The prompting guidelines (reproduced in Figure \ref{fig:assumptions-prompt}) also enforce several quality constraints: the database context is designated as authoritative, whereas the web context is treated as supplementary background only. Furthermore, at least one assumption must be grounded in a specific triple from the database context, cited by its sequence number. Assumptions are prioritized by semantic value, i.e., assumptions about what the entity is and how it is used in practice are preferred over structural schema axioms, which are only included when they reveal something semantically distinctive. Assumptions supported by instance-level data are considered the most valuable.

While structuring intermediate assumptions reduces noise and improves factual traceability, it introduces a theoretical risk of information loss. Because this step acts as a semantic bottleneck between raw retrieval and final generation, any crucial detail omitted by the model during the synthesis phase becomes permanently unavailable to the description generation agent, potentially reducing the completeness of the final output. To empirically evaluate this trade-off between noise reduction and over-summarization, Section~\ref{sec:experiments} reports an ablation study comparing descriptions generated with and without the assumption generation step. In the ablation condition, the description agent receives the raw database and web contexts directly, bypassing assumption generation entirely.

\subsection{Description Generation} \label{sec:des_generation}

The final stage of the pipeline produces the natural language description. The description generation agent is prompted as a professional lexicographer (see Figure \ref{fig:description-prompt}) and is given the structured assumption list from the previous stage (or, in the ablation condition, the raw contexts). 
The agent is instructed to write a concise, accurate description that
(i)~does not name the IRI directly; (ii)~uses specific terminology
derived from the database context assumptions where relevant;
(iii)~omits generic metadata properties (identifiers, dates, status
flags) unless they are the defining characteristic of the entity; and
(iv)~favours rare, distinctive facts that differentiate the entity from
similar concepts in the same ontology. The following examples from the Friend of a Friend ontology are
provided in the prompt to calibrate expected output style:

\texttt{membershipClass} $\to$ \emph{Indicates the class of
    individuals that are a member of a Group.}

\texttt{Agent} $\to$ \emph{A resource that acts or has the power
    to act.}

Once the description text is composed, the agent calls a \texttt{persist\_description} tool, which logs the description together with the pipeline configuration metadata (model names and ablation flags) to a database. This logging serves two purposes: it enables post-hoc experiment tracking and supports the ablation analysis reported in Section~\ref{sec:experiments}. The agent's final response consists solely of the description text, which is written to the shared pipeline under the key \texttt{description} and returned as the output of the full pipeline.

%% file: sec-experiments.tex
\section{Experiments and Evaluation}
\label{sec:experiments}

We evaluate OntoDoc through a sequence of experiments designed to answer four questions: 1) \emph{Which underlying LLM provides the most reliable performance within the OntoDoc pipeline?} 2) \emph{Does the intermediate assumption generation stage improve description quality?} 3) \emph{How does OntoDoc compare against existing ontology description generation baselines?} 4) \emph{Do automatic evaluation trends align with human judgement?}  To answer these questions, we conduct a model selection study, an ablation study on the assumption generation component, automatic evaluation against baseline systems, and human evaluation using domain experts and student annotators.

\subsection{Experimental Setup}

{\flushleft \textbf{Datasets}.}\hspace{2mm} 
We evaluate OntoDoc on 151 ontology terms (each an IRI) drawn from three knowledge graphs, covering diverse domains and ontology structures across biomedical, agricultural, and open-domain settings.

\textbf{BioBricks-MeSH} \cite{biobricks_mesh} (51 IRIs): A biomedical knowledge graph derived from the Medical Subject Headings (MeSH) vocabulary. The ontology contains 17 classes, 21 object properties, and 28 datatype properties. 
    
\textbf{SOCKG} \cite{shirvani2025knowledge} (70 IRIs identified as semantically important by graph developers and soil scientists): A domain-specific agricultural knowledge graph developed for soil science research using the USDA GRACEnet (Greenhouse gas Reduction through Agricultural Carbon Enhancement Network) dataset. The ontology contains 92 classes, 62 object properties, and 10 datatype properties. 
    
\textbf{DBpedia} \cite{dbpedia} (30 randomly sampled IRIs): A large-scale open-domain knowledge graph automatically extracted from Wikipedia. The ontology contains 790 classes, 1167 object properties, and 1857 datatype properties.

{\flushleft \textbf{Evaluation metrics}.}\hspace{2mm} 
We evaluate generated descriptions using four quality dimensions derived from the intended purpose of \texttt{rdfs:comment} annotations (full scoring rubrics are provided in Tables \ref{tab:rubric-accuracy}--\ref{tab:rubric-cruciality} in the Annex):

\emph{Accuracy}: whether the description correctly represents the concept.

\emph{Clarity}: whether the description is readable and accessible to non-experts.

\emph{Completeness}: whether the description captures the important semantic information necessary to understand the term.
 
\emph{Cruciality}: whether the included information is essential and non-redundant.

{\flushleft \textbf{Reference descriptions}.}\hspace{2mm} 
Although DBpedia and Biobricks MeSH contain existing \texttt{rdfs:comment} annotations for many terms, we do not treat these annotations as gold-standard reference descriptions. During preliminary inspection, we observed that many comments function primarily as implementation notes or partial usage hints rather than semantically complete descriptions. In some cases, annotations also contradicted information explicitly present in the knowledge graph itself. Instead, evaluation is performed directly on generated descriptions using the four quality dimensions above, with the verified database context serving as the shared factual grounding for both automatic and human judges.

{\flushleft \textbf{Grounding the descriptions}.}\hspace{2mm} 
A key challenge is that the database context is produced via an agentic retrieval process, where outputs depend on model-selected queries. Consequently, retrieved contexts are nondeterministic and may vary across models and runs, introducing potential evaluation bias. To ensure fairness, we standardize this step by running the database context retrieval step as outlined in Section~\ref{sec:db_context} using the same model that serves as the LLM judge. We refer to the resulting database contexts outputs as \emph{canonical database contexts}. Using this setup, for each term in the combined dataset, we construct a single context represented as a numbered list of RDF triples, optionally including instance counts. To ensure factual correctness, we apply an automated cleaning and verification pipeline to each canonical database context. Each context is then automatically cleaned: triples are validated against the knowledge graph via SPARQL ASK queries, instance counts are verified for consistency, and any hallucinated or inconsistent information is discarded, yielding a fully verified canonical context.

\subsection{Model Selection and Ablation Study}

Two preliminary experiments are conducted in a deliberate sequence: model selection first, followed by the ablation study on assumption generation. This ordering is motivated by the need to isolate the contribution of each factor independently.

{\flushleft \textbf{Model selection}.}\hspace{2mm} 
We first evaluate which underlying language model provides the strongest performance within the OntoDoc pipeline. We compare three candidate generation models: GPT-4.1 Mini \cite{openai2025gpt41mini}, Gemini 2.5 Flash Lite \cite{google2025gemini25flashlite}, and GLM-4.7 Flash \cite{zhipu2026glm47flash}. Evaluation is conducted using an LLM judge (Kimi K2.5 \cite{bai2026kimik25}) in a forced-choice preference setting rather than rubric-based scoring. For each IRI, OntoDoc generates two descriptions per candidate model (one description generated using the assumptions step, and one description without). The judge is then presented with the target IRI, the canonical database context, and the three candidate descriptions in randomized order. The judge is then instructed to select the single best description according to the evaluation criteria of accuracy, clarity, completeness, and cruciality in a forced-choice preference setting (see the prompt template in Figure \ref{fig:model-selection-prompt}). We repeat this step using the descriptions that were generated without the assumptions step.

Thus, for each ontology term, two independent comparisons are performed: one among three descriptions generated with assumptions (one per model), and one among three descriptions generated without assumptions. Table~\ref{tab:model_selection} reports model win rates with and without the assumption generation stage. Across both conditions, GPT-4.1 Mini achieves the highest and most stable win rate, so we use it as the default for all remaining experiments.

\begin{table}[t]
    \centering
    \caption{Win Rates with and without Assumptions}
    \resizebox{\columnwidth}{!}{%
    \begin{tabular}{|l|c|c|}
        \hline
        \textbf{Model} & \textbf{Win Rate (w/ assumptions)} & \textbf{Win Rate (w/o assumptions)} \\
        \hline
        gpt-4.1-mini & 47.02\% & 47.68\% \\
        \hline
        gemini-2.5-flash-lite & 23.18\% & 19.21\% \\
        \hline
        glm-4.7-flash & 29.80\% & 33.11\% \\
        \hline
    \end{tabular}}
    \label{tab:model_selection}
\end{table}

{\flushleft \textbf{Assumption generation ablation study}.}\hspace{2mm} 
We next evaluate the contribution of the assumption generation component. Using \texttt{gpt-4.1-mini} as the fixed generation model, we compare OntoDoc with and without the assumption generation stage. The evaluation procedure mirrors the model selection setup. For each IRI, the judge receives two candidate descriptions generated under the two configurations together with the canonical database context and selects the stronger description using the ablation prompt template (Figure \ref{fig:ablation-prompt}). Table~\ref{tab:kg_assumptions} reports the percentage of cases in which the configuration with assumption generation was preferred. The assumption generation stage improves performance consistently across all three knowledge graphs. The largest gains are observed on BioBricks MeSH, suggesting that explicitly reasoning over retrieved evidence is particularly beneficial in highly specialized scientific domains. Based on these results, the assumption generation stage is included in the final OntoDoc configuration.

\vspace{-0.3 cm}
\begin{table}[ht]
    \centering
    \caption{Assumptions Step Win Rate per Knowledge Graph for gpt-4.1-mini}
    \begin{tabular}{|l|c|c|}
        \hline
        \textbf{Knowledge Graph} & \textbf{Sample Size (n)} & \textbf{Assumptions Step Win Rate} \\
        \hline
        DBpedia & 30 & 57.00\% \\
        \hline
        BioBricks-MeSH & 51 & 61.00\% \\
        \hline
        SOCKG & 70 & 54.00\% \\
        \hline
    \end{tabular}
    \label{tab:kg_assumptions}
\end{table}
\vspace{-0.3 cm}

\subsection{Automatic Evaluation}

For end-to-end evaluation, we compare OntoDoc against two baseline systems:

\textbf{Single-Hop Retrieval (Naive).} This baseline simulates the behavior of a user unfamiliar with the knowledge graph who issues a simple outgoing-triple query: \texttt{
SELECT ?p ?o WHERE \{ <iri> ?p ?o . \}. The exact LLM prompt that was given to gpt-4.1-mini for generating these descriptions is shown in Figure~\ref{fig:naive-prompt}.
}
The retrieved triples are passed directly to an LLM without additional graph exploration, reasoning, or web retrieval.

\textbf{Ontology Verbalizer.} A schema-based verbalization system that generates natural language descriptions from ontology axioms and logical structure.

Because the ontology verbalizer only supports IRIs that are explicitly typed as \texttt{owl:Class}, direct comparison across all three methods is restricted to a subset of ontology terms supported by every approach. We refer to this subset as the \emph{common IRI set}.

{\flushleft \textbf{Comparison with naive baseline}.}\hspace{2mm} 
We first evaluate OntoDoc against the naive single-hop baseline across all 151 terms using \texttt{kimi-k2.5} as the LLM judge. Each description is independently scored on a 5-point Likert scale for accuracy, clarity, completeness, and cruciality by the LLM judge (see Figure \ref{fig:method-comparison-prompt} for the grading prompt configuration). Table~\ref{tab:ontodoc-naive} presents the per–knowledge graph mean scores for both systems. OntoDoc consistently achieves substantially higher completeness across all three graphs: on DBpedia, completeness rises from 3.67 (naive) to 3.97; on BioBricks MeSH from 3.37 to 3.71; and on SOCKG from 3.26 to 3.84. Averaged across all knowledge graphs, OntoDoc attains a mean completeness of 3.84 compared to 3.43 for the naive baseline, confirming that the multi‑stage evidence synthesis produces significantly more thorough descriptions.

Accuracy and clarity follow a slightly different pattern. On DBpedia, the naive baseline yields marginally higher accuracy (4.30 vs.\ 4.20) and clarity (4.73 vs.\ 4.47), likely because the simple triples it retrieves are easier to paraphrase without introducing subtle distortions. On the more specialised BioBricks MeSH and SOCKG graphs, however, OntoDoc matches or exceeds the naive baseline in both dimensions (e.g., accuracy 3.76 vs.\ 3.65 on BioBricks MeSH; 3.91 vs.\ 3.90 on SOCKG), demonstrating that its adaptive exploration becomes increasingly beneficial as domain complexity grows. Cruciality scores are similar across systems, with OntoDoc maintaining high conciseness (3.79–4.04) while incorporating richer content. Overall, the automatic evaluation shows that OntoDoc delivers a clear advantage in completeness without sacrificing the other quality dimensions, and in domain‑specific settings it also equals or improves upon the naive baseline in accuracy and clarity.

\begin{table}[htbp]
\centering
\caption{OntoDoc vs Naive Scores per Knowledge Graph}
\begin{tabular}{llcc}
\hline
\textbf{Knowledge Graph} & \textbf{Metric} & \textbf{OntoDoc} & \textbf{Naive} \\
\hline
\multirow{4}{*}{DBpedia (n=30)} & Accuracy & 4.20 & 4.30 \\
 & Clarity & 4.47 & 4.73 \\
 & Completeness & 3.97 & 3.67 \\
 & Cruciality & 3.90 & 4.43 \\
\hline
\multirow{4}{*}{BioBricks-MeSH (n=51)} & Accuracy & 3.76 & 3.65 \\
 & Clarity & 4.43 & 4.24 \\
 & Completeness & 3.71 & 3.37 \\
 & Cruciality & 4.04 & 4.00 \\
\hline
\multirow{4}{*}{SOCKG (n=70)} & Accuracy & 3.91 & 3.90 \\
 & Clarity & 4.49 & 4.37 \\
 & Completeness & 3.84 & 3.26 \\
 & Cruciality & 3.79 & 3.97 \\
\hline
\end{tabular}
\label{tab:ontodoc-naive}
\end{table}

{\flushleft \textbf{Comparison with ontology verbalizer}.}\hspace{2mm} 
For the common IRI subset (those unambiguously typed as \texttt{owl:Class}), we additionally compare OntoDoc with the template-based ontology verbalizer. The same LLM judge and scoring protocol are applied. The automatic scores per knowledge graph is shown in Table~\ref{tab:auto-score-per-kg}. In terms of accuracy, OntoDoc either matches or slightly exceeds the verbalizer across all three knowledge graphs. The most pronounced difference lies once again in completeness, where OntoDoc vastly outperforms the verbalizer, achieving scores that are often one full point higher on the 5‑point scale. This confirms that the template‑driven verbalizer, limited to explicit schema axioms, misses much of the semantic information that OntoDoc recovers from instance data and external knowledge. Clarity and cruciality scores are comparable, with OntoDoc maintaining high readability and conciseness despite its richer content.

\begin{table}[htbp]
\centering
\caption{Rubric Scores (common URIs) per Knowledge Graph}
\begin{tabular}{llccc}
\hline
\textbf{Knowledge Graph} & \textbf{Metric} & \textbf{Naive} & \textbf{OntoDoc} & \textbf{Verbalizer} \\
\hline
\multirow{4}{*}{DBpedia} & Accuracy & 4.20 & 4.30 & 1.50 \\
 & Clarity & 4.80 & 4.70 & 1.00 \\
 & Completeness & 3.40 & 4.00 & 1.20 \\
 & Cruciality & 4.50 & 4.10 & 1.50 \\
\hline
\multirow{4}{*}{BioBricks-MeSH} & Accuracy & 3.65 & 4.00 & 3.41 \\
 & Clarity & 4.24 & 4.71 & 3.94 \\
 & Completeness & 3.24 & 3.88 & 3.18 \\
 & Cruciality & 4.12 & 4.35 & 3.65 \\
\hline
\multirow{4}{*}{SOCKG} & Accuracy & 4.00 & 4.12 & 3.96 \\
 & Clarity & 4.42 & 4.58 & 4.58 \\
 & Completeness & 3.33 & 3.96 & 1.75 \\
 & Cruciality & 3.96 & 3.75 & 4.96 \\
\hline
\end{tabular}
\label{tab:auto-score-per-kg}
\end{table}
\vspace{-0.5 cm}

\subsection{Human Evaluation}

To validate the quality of OntoDoc-generated descriptions in a realistic setting, we conduct a human evaluation study. Unlike the automatic evaluation, which compares OntoDoc against baseline systems, the human evaluation focuses exclusively on assessing the quality of OntoDoc outputs in isolation, providing an absolute measure of description quality from the perspective of real users.

For SOCKG, annotations are performed by external domain experts in soil science. For DBpedia and BioBricks MeSH, annotations are performed by student evaluators with prior experience working with RDF knowledge graphs. In total, 11 annotators participated in the study. For each ontology term, annotators are presented with the target IRI, the canonical database context, and the generated description. For SOCKG, a screenshot of the relevant section of the AgCROS dataset that the term is modeled after is also provided to aid the domain experts. Descriptions are evaluated independently according to the four evaluation dimensions using the same 5-point Likert scale employed in the automatic evaluation.

Table ~\ref{tab:human-eval} summarizes the average human evaluation scores across datasets. Human evaluators rated OntoDoc descriptions highly across all dimensions and knowledge graphs. On DBpedia and BioBricks MeSH, mean scores for clarity and accuracy comfortably exceed 4.0 on the 5‑point scale, while completeness and cruciality also receive strongly positive ratings. The pattern mirrors the automatic evaluation: clarity and accuracy tend to be the highest‑scored dimensions, followed by cruciality and completeness, which reflects the inherent difficulty of packing all essential information into a concise description without introducing any extraneous detail. Scores on SOCKG are slightly lower overall, consistent with the increased domain specificity and terminological complexity of agricultural and soil science concepts; nonetheless, all mean ratings remain in the favourable range, demonstrating that OntoDoc produces high‑quality descriptions even for highly specialised scientific terms.

The alignment between human and automatic evaluation trends confirms that the LLM‑based judge captures the same quality perceptions as human evaluators, supporting the reliability of the automatic scoring protocol used in the preceding experiments.

\vspace{-0.3 cm}
\begin{table}[htbp]
\centering
\caption{Average Human Annotation Scores by Knowledge Graph}
\begin{tabular}{lcccc}
\hline
\textbf{KG} & \textbf{Accuracy} & \textbf{Clarity} & \textbf{Completeness} & \textbf{Cruciality} \\
\hline
DBpedia & 4.43 & 4.36 & 4.36 & 4.20 \\
BioBricks-MeSH & 4.51 & 4.49 & 4.43 & 4.34 \\
SOCKG & 3.93 & 3.65 & 3.72 & 3.74 \\
\hline
\end{tabular}
\label{tab:human-eval}
\end{table}
\vspace{-0.5 cm}

%% file: sec-results.tex
\section{Discussion}

Our experiments reveal a consistent pattern: OntoDoc's primary advantage lies in the completeness of its generated descriptions. Adaptive graph exploration combined with external contextual grounding allows it to recover semantic information that fixed-query and schema-only approaches miss. This is most visible in comparisons against the ontology verbalizer, whose reliance on explicit schema axioms leaves it unable to surface semantics expressed implicitly through instance data.

These gains carry a trade-off. The cruciality metric rewards descriptions in which every detail is strictly necessary and since OntoDoc retrieves substantially more evidence than the baselines, it is more likely to include secondary details that are informative but not absolutely essential. This is particularly visible on SOCKG, where the verbalizer achieves high cruciality despite poor completeness due to its tendency to produce structurally generic outputs (e.g., ``Planting management is a type of measurable entity.'') that contain little opportunity for irrelevant content. High cruciality in this case reflects brevity, not quality.

The DBpedia results point to a boundary condition: when concepts are already well represented in a model's pretraining data, the additional context OntoDoc gathers offers marginal benefit and can occasionally introduce detail that slightly harms clarity or cruciality. OntoDoc is most valuable in specialized domains where underlying model knowledge is sparse.

Finally, the lower SOCKG scores in human evaluation reflect annotator composition as much as system performance. Domain experts in agricultural soil science apply stricter standards of scientific precision than general RDF-experienced annotators, so the gap between SOCKG and the other datasets is best interpreted as a difficulty effect rather than a system failure. The alignment between human and automatic evaluation trends across all three graphs confirms that LLM-based scoring reliably tracks human quality perception.

%% file: sec-conclusion.tex
\section{Conclusion}
We presented OntoDoc, a multi-agent framework for automatically generating natural language descriptions for ontology terms in heterogeneous knowledge graphs. By combining adaptive SPARQL-based graph exploration, external web retrieval, and an intermediate assumption generation stage, OntoDoc produces descriptions that are substantially more complete than naive single-hop retrieval or template-based ontology verbalization, while maintaining competitive accuracy and clarity across biomedical, agricultural, and open-domain settings.
Our evaluation across 151 ontology terms demonstrates that the assumption generation stage consistently improves description quality, particularly in specialized scientific domains. Human evaluation by both domain experts and student annotators corroborates the trends observed in automatic scoring, confirming the reliability of LLM-based assessment for this task.

%% file: sec-genai.tex
\section{Declaration of use of Generative AI}
We used generative AI to improve the clarity, grammar, and readability of portions of the text. All AI-suggested changes were reviewed and finalized by the authors. In addition, we used Claude to assist in writing and refactoring sections of the code used in this research. The authors verified, tested, and assume full responsibility for all generated code.

No generative AI tool is listed as a co-author, nor was any AI used to create figures, tables, or citations. All use described above complies with the conference's disclosure requirements, and the authors remain fully responsible for the originality, accuracy, and integrity of the entire work.

%% file: appendix.tex
\section{Annex}

\FloatBarrier

\begin{tcblisting}{
    colback=white,
    colframe=black,
    title=\ttfamily Database Context Agent System Prompt,
    fonttitle=\bfseries,
    boxrule=0.5pt,
    arc=1mm,
    left=2mm,
    right=2mm,
    top=1mm,
    bottom=1mm,
    enhanced,
    breakable,
    listing only,
    listing options={
        basicstyle=\ttfamily\small,
        breaklines=true,
        breakatwhitespace=true,
        columns=fullflexible
    }
}
You are a database context agent for the ONTODOC pipeline. Your goal is to gather
enough context about the target IRI from the knowledge graph so that the downstream
assumptions agent can write an accurate rdfs:description.

IRI:              {iri}
SPARQL endpoint:  {sparql_endpoint}
Triple budget:    {max_triples}  (stop exploring once you have accumulated this many triples)
Tool-call budget: {max_tool_calls} (stop exploring once you have made this many tool calls)

You have the following tools available:

  get_iri_information   - rdf:type, label, comment, description, definition for any IRI,
                          plus whether it appears as a predicate (i.e. is a property).
  get_outgoing_triples  - triples where the IRI is the subject:  <iri> ?p ?o
  get_incoming_triples  - triples where the IRI is the object:   ?s ?p <iri>
  get_predicate_usage   - sample triples where the IRI is the predicate: ?s <iri> ?o,
                          plus distinct subject and object types. Only useful for properties.
  execute_sparql        - arbitrary SPARQL SELECT or ASK for anything the above cannot express.

---

STOPPING CONDITIONS - check after every tool call.
Stop exploring and go directly to Phase 4 when ANY of these are met:
  - You have accumulated {max_triples} triples or more across all results.
  - You have made {max_tool_calls} tool calls in total.
  - You have characterised all significant IRIs and further exploration is clearly redundant.

---

PHASE 1 - Orient

1. Call get_iri_information on the target IRI.

From the results, determine the IRI type:
  - CLASS:    rdf:type is owl:Class or rdfs:Class, OR many incoming rdf:type triples.
  - PROPERTY: get_iri_information reports "Is property: Yes", OR rdf:type is
              owl:ObjectProperty or owl:DatatypeProperty.
  - INSTANCE: rdf:type points to a domain class (not owl:Class / rdfs:Class).
  - UNKNOWN:  None of the above.

---

PHASE 2 - Initial scan

Make the following tool calls based on the IRI type determined in Phase 1.

  CLASS:
    - get_incoming_triples(limit=20)  - find instances and subclasses.
    - get_outgoing_triples(limit=20)  - find metadata, superclasses, restrictions.
    - execute_sparql: COUNT(*) of instances of this class.
      SELECT (COUNT(?i) AS ?n) WHERE { ?i a <{iri}> . }
      If the count is small (< 10), also retrieve all instances explicitly.

    IMPORTANT: The incoming triples will mostly be rdf:type assertions from instances.
    These instance IRIs themselves are not informative - what matters is what those
    instances look like. Immediately pick 2-3 distinct instance IRIs from the results
    and call get_outgoing_triples on each. This reveals the actual attributes and
    relationships that instances of this class carry, which is the most valuable
    information for describing the class.

  PROPERTY:
    - get_predicate_usage(limit=20)   - find what subjects and objects it connects.
    - get_outgoing_triples(limit=20)  - find domain, range, and other metadata.

    Then pick one subject IRI and one object IRI from the usage results and call
    get_outgoing_triples on each to understand the entities this property connects.

  INSTANCE:
    - get_outgoing_triples(limit=20)  - find attribute values.
    - get_incoming_triples(limit=10)  - find what references this instance.

  UNKNOWN:
    - get_outgoing_triples(limit=10)
    - get_incoming_triples(limit=10)

---

PHASE 3 - Targeted follow-up

Examine all results gathered so far. For each IRI that looks semantically significant
- a class connected to the target, a property that appears on instances, an object
that multiple instances point to - do the following, always subject to the stopping
conditions above:

1. Call get_iri_information.
2. If it is a class: call get_outgoing_triples(limit=10) on 1-2 of its instances
   to see what attributes those instances carry.
3. If it is a property you have not yet explored:
     call get_predicate_usage(limit=10).
4. Use execute_sparql for targeted questions, e.g.:
     - DISTINCT values of a datatype property across instances.
     - COUNT of instances of a class.
     - All distinct predicates used by instances of a class:
         SELECT DISTINCT ?p WHERE { ?s a <class> . ?s ?p ?o . }

The goal is to understand the full shape of the data: what attributes instances
carry, what other classes they connect to, and what the values look like.
Do NOT explore pure schema IRIs: owl:Class, rdfs:Class, rdf:Property,
owl:Ontology, or XSD datatypes.

---

PHASE 4 - Summarize

Compile the most informative triples gathered across all phases into a single
numbered list using exactly this format:

  1. <subject> <predicate> <object> .
  2. <subject> <predicate> <object> .
  ...

Rules:
  - Number triples sequentially starting from 1.
  - DEDUPLICATION: When many instances share the same structure (e.g. 20 triples
    that all say ?x rdf:type SomeClass), include at most 2-3 as representatives.
    The assumptions agent does not benefit from seeing the same pattern repeated.
  - PRIORITIZATION: Prefer triples that reveal relationships between classes,
    attribute values on instances, and connections to other named entities over
    bare rdf:type assertions. A triple like:
      MeasSoilChem.10 sockg:fromUnit ExperimentalUnit.5 .
    is far more informative than:
      MeasSoilChem.10 rdf:type SoilChemicalSample .
  - Omit pure schema/infrastructure triples:
      rdf:type owl:Class, rdf:type rdfs:Class, rdf:type owl:Ontology,
      rdf:type owl:ObjectProperty, rdf:type owl:DatatypeProperty,
      owl:imports, owl:versionIRI, rdf:type owl:NamedIndividual.
  - Omit exact duplicates.
  - If the COUNT query returned a small number (<10), ensure that all rdf:type triples are listed in this summary. Then, after the end of the list, in a new line, write the following sentence: "There are exactly N instances and they are all listed above."
  - If the COUNT query returned a large number (>= 10), in a new line after the end of the list, write the following sentence: "There are exactly N instances."

Respond with ONLY the numbered triple list. No headings, no explanations,
no follow-up questions.
\end{tcblisting}

\captionof{figure}{System Prompt for the Database Context Agent}
\label{fig:db-context-prompt}

\FloatBarrier

\begin{tcblisting}{
    colback=white,
    colframe=black,
    title=\ttfamily Web Context Agent System Prompt,
    fonttitle=\bfseries,
    boxrule=0.5pt,
    arc=1mm,
    left=2mm,
    right=2mm,
    top=1mm,
    bottom=1mm,
    enhanced,
    breakable,
    listing only,
    listing options={
        basicstyle=\ttfamily\small,
        breaklines=true,
        breakatwhitespace=true,
        columns=fullflexible
    }
}
You are a web context agent for the ONTODOC pipeline. Your goal is to gather external
encyclopedic or technical context for an ontology IRI to aid in generating an accurate
rdfs:description.

IRI: {iri}

Steps:
1. Extract the label from the IRI - the fragment after the last '/' or '#'.
2. If the label is a trivial, unambiguous common English word (e.g. "name", "id", "date",
   "type", "value", "label", "title", "description", "comment"), respond with an empty
   string immediately. Do NOT call any tools.
3. Otherwise, determine the complexity of the label and choose the appropriate tool(s):
   - **fetch_wikipedia_context**: Use this for general-purpose, common, or academic terms
     that are likely to have a dedicated Wikipedia entry (e.g. "Nitrite", "Agent", "Policy").
   - **fetch_tavily_context**: Use this for highly technical, niche, or specific terms
     that might not have a Wikipedia entry but exist in technical documentation or web
     resources (e.g. "nitritesSd_mgN_per_kg").
   
   You may call one or both tools if the label is compound or ambiguous. Ensure you
   decompose the label into simple, searchable keywords first.
4. Respond with the combined web context text returned by the tools, or an empty string if
   no context was found.

Do not ask follow-up questions at the end of your response.
\end{tcblisting}

\captionof{figure}{System Prompt for the Web Context Agent}
\label{fig:web-context-prompt}

\FloatBarrier

\begin{tcblisting}{
    colback=white,
    colframe=black,
    title=\ttfamily Assumptions Agent System Prompt,
    fonttitle=\bfseries,
    boxrule=0.5pt,
    arc=1mm,
    left=2mm,
    right=2mm,
    top=1mm,
    bottom=1mm,
    enhanced,
    breakable,
    listing only,
    listing options={
        basicstyle=\ttfamily\small,
        breaklines=true,
        breakatwhitespace=true,
        columns=fullflexible
    }
}
You are a professional lexicographer extracting structured assumptions about a scientific
or technical IRI. Your goal is to produce assumptions in JSON format that will be used
to write a clear, accurate rdfs:description.

IRI: {iri}
Database Context: {database_context}
Web Context: {web_context}

You have two sources of information:
1. **Database Context** - specific technical details. Treat as authoritative.
2. **Web Context** - general Wikipedia-based overview. Treat as unreliable reference only.

First, determine whether the IRI is a CLASS, PROPERTY, or INSTANCE by inspecting the
Database Context (rdf:type, whether it appears as a predicate, etc.).
Then follow the corresponding guidance below.

IF THE IRI IS A CLASS:

- Prefer assumptions about WHAT the entity is and HOW IT IS USED over structural schema facts.
- Schema facts (subClassOf, disjointWith, domain/range) should only appear when they reveal.
  something semantically distinctive that would not be obvious otherwise.
- Assumptions grounded in usage annotations or instance-level data are more valuable than pure schema axioms.
- If the COUNT query returned a small number and all instances are explicitly listed in the Database Context, state the exact count as a definitive fact, not a hedged observation.
  "There are exactly N instances" not "the sample contains N instances".
- If there are 3 instances or less, state "x, y, and z are the only instances of this class."
- If there are more than 3 instances, state "instances include but are not limited to x, y, ans z."

IF THE IRI IS A PROPERTY:

Focus on these questions in order of priority:

1. **What do the values mean semantically?**
   Do not just say "a string" or "a number." Look at the example values in the Database
   Context and reason about what they represent in the real world (e.g. CAS numbers,
   percentages, ISO dates, controlled vocabulary terms). Use the Web Context to help
   identify value semantics if available.

2. **What entities use this property, and in what context?**
   State the domain class and, where possible, the broader context in which the property
   is meaningful (e.g. "used for Concepts related to Drugs and Chemicals").

3. **What is the cardinality pattern?**
   Check whether multiple values appear on a single subject in the Database Context.
   If so, explicitly state that multiple values are allowed and explain why that makes
   sense (e.g. salts, stereoisomers, aliases).

4. **How does this property relate to similar properties?**
   If the database context reveals related properties, note what distinguishes
   this one (e.g. "unlike registryNumber, this property captures additional identifiers").
  
For each assumption:
- Provide your reasoning in the "reasoning" field.
- State the assumption clearly in the "assumption" field.
- For assumptions grounded in Database Context, cite the relevant triple number(s).
- Ensure at least one assumption is grounded directly in the Database Context.

Respond only with a JSON object in this exact format:
{
  "items": [
    {
      "reasoning": "reasoning here",
      "assumption": "assumption here"
    }
  ]
}

Do not ask follow-up questions at the end of your response.
\end{tcblisting}

\captionof{figure}{System Prompt for the Assumptions Agent}
\label{fig:assumptions-prompt}

\FloatBarrier

\begin{tcblisting}{
    colback=white,
    colframe=black,
    title=\ttfamily Description Agent System Prompt,
    fonttitle=\bfseries,
    boxrule=0.5pt,
    arc=1mm,
    left=2mm,
    right=2mm,
    top=1mm,
    bottom=1mm,
    enhanced,
    breakable,
    listing only,
    listing options={
        basicstyle=\ttfamily\small,
        breaklines=true,
        breakatwhitespace=true,
        columns=fullflexible
    }
}
You are a professional lexicographer writing a natural language rdfs:description for an
ontology IRI.

IRI: {iri}
Assumptions: {assumptions}

Guidelines:
- Write 1-3 sentences maximum. If you find yourself writing more, cut.
- Treat the Database Context as authoritative; use the Web Context only as supplementary background.
- Do not name the IRI directly.
- Describe WHAT the entity is and HOW it is used and not its schema structure or metadata
- Be informative and accessible to someone unfamiliar with the ontology.
- Use specific terminology from the Database Context assumptions where relevant.
- Lead with what makes this entity unusual or distinctive compared to similar entities in the same ontology.
- Generic structural facts (disjointWith, preferredConcept, versioning dates) should be omitted entirely unless no more distinctive facts exist.
- If you decide to mention example IRIs, keep it to at most 3 and do not write the full IRI (e.g., write D000001 instead of http://id.nlm.nih.gov/mesh/D000001)
- Do not mention the number of instances unless the count is the definining characteristic of the class or the number if instances is notably small (less than 10)
- Output only the description text - no explanations, formatting, or reasoning.
- Do not ask follow-up questions at the end of your response.

Bad example (too long, schema-dumping):
  TopicalDescriptor - "Each instance carries descriptor metadata - identifier, active status,
  dateCreated/dateRevised, historyNote and nlmClassificationNumber - and is linked to concepts,
  preferred terms, tree numbers..."

Good examples:
  membershipClass  - Indicates the class of individuals that are a member of a Group
  Policy           - A plan or course of action by an authority, intended to influence and
                     determine decisions, actions, and other matters
  Agent            - A resource that acts or has the power to act

After generating the description, call persist_description with:
  iri="{iri}"
  description=<your generated description>
  sparql_endpoint="{sparql_endpoint}"

Then respond with only the description text.
\end{tcblisting}
\captionof{figure}{System Prompt for the Description Agent}
\label{fig:description-prompt}

\FloatBarrier

\begin{tcblisting}{
    colback=white,
    colframe=black,
    title=\ttfamily Model Selection Prompt Template,
    fonttitle=\bfseries,
    boxrule=0.5pt,
    arc=1mm,
    left=2mm,
    right=2mm,
    top=1mm,
    bottom=1mm,
    enhanced,
    breakable,        
    listing only,      
    listing options={
        basicstyle=\ttfamily\small,
        breaklines=true,
        breakatwhitespace=true,
        columns=fullflexible
    }
}
You are an expert knowledge graph curator evaluating the quality of automatically generated \
descriptions for ontology IRIs.

You will be shown:
  1. A database context - a numbered list of RDF triples that describe the IRI and its relationships \
in the knowledge graph. Treat this as a sampled reference, but note that it is not necessarily the complete information about the IRI.
  2. Three candidate descriptions generated by different AI systems.

Your task is to select the single best description based on these criteria:
  1. Accuracy             - does the description correctly reflect what can be expected of the IRI?
  2. Completeness         - does it capture all key information about the IRI?
  3. Cruciality           - Is all of the included information essential for understanding the concept?
  4. Clarity              - is it readable and accessible to someone unfamiliar with the ontology?
---

IRI: {iri}

DATABASE CONTEXT:
{context}

---

CANDIDATE DESCRIPTIONS:
1. {desc_1}
2. {desc_2}
3. {desc_3}

---

Respond with a JSON object and nothing else - no explanation, no markdown fences:
{{"winner": <1, 2, or 3>}}
\end{tcblisting}
\captionof{figure}{Prompt Template for Model Selection}
\label{fig:model-selection-prompt}

\FloatBarrier

\begin{tcblisting}{
    colback=white,
    colframe=black,
    title=\ttfamily Ablation Study Prompt Template,
    fonttitle=\bfseries,
    boxrule=0.5pt,
    arc=1mm,
    left=2mm,
    right=2mm,
    top=1mm,
    bottom=1mm,
    enhanced,
    breakable,
    listing only,     
    listing options={
        basicstyle=\ttfamily\small,
        breaklines=true,
        breakatwhitespace=true,
        columns=fullflexible
    }
}
You are an expert knowledge graph curator evaluating the quality of automatically generated \
descriptions for ontology IRIs.

You will be shown:
  1. A database context - a numbered list of RDF triples that describe the IRI and its relationships \
in the knowledge graph. Treat this as a sampled reference, but note that it is not necessarily the complete information about the IRI.
  2. Two candidate descriptions (A and B) generated by the same AI system under different conditions.

Your task is to select the single better description based on these criteria:
  1. Accuracy             - does the description correctly reflect what can be expected of the IRI?
  2. Completeness         - does it capture all key information about the IRI?
  3. Cruciality           - Is all of the included information essential for understanding the concept?
  4. Clarity              - is it readable and accessible to someone unfamiliar with the ontology?
---

IRI: {iri}

DATABASE CONTEXT:
{context}

---

CANDIDATE DESCRIPTIONS:
A: {desc_a}
B. {desc_b}

---

Respond with a JSON object and nothing else - no explanation, no markdown fences:
{{"winner": "A" or "B"}}
\end{tcblisting}
\captionof{figure}{Prompt Template for the Ablation Study}
\label{fig:ablation-prompt}

\FloatBarrier

\begin{tcblisting}{
    colback=white,
    colframe=black,
    title=\ttfamily Naive Generation Prompt Template,
    fonttitle=\bfseries,
    boxrule=0.5pt,
    arc=1mm,
    left=2mm,
    right=2mm,
    top=1mm,
    bottom=1mm,
    enhanced,
    breakable,
    listing only,
    listing options={
        basicstyle=\ttfamily\small,
        breaklines=true,
        breakatwhitespace=true,
        columns=fullflexible
    }
}
Your task is to write a brief natural language description for an ontology IRI.

IRI: {iri}
Knowledge Graph: {kg_name}

The following triples describe what is known about this IRI in the knowledge graph:
{triples}

Guidelines:
- Write 1-3 sentences maximum.
- Do not name the IRI directly.
- Be informative and accessible to someone unfamiliar with the ontology.
- Output only the description text - no explanations or formatting.
\end{tcblisting}
\captionof{figure}{Prompt Template for the Naive Generation Baseline}
\label{fig:naive-prompt}

\FloatBarrier

\begin{tcblisting}{
    colback=white,
    colframe=black,
    title=\ttfamily Method Comparison Prompt Template,
    fonttitle=\bfseries,
    boxrule=0.5pt,
    arc=1mm,
    left=2mm,
    right=2mm,
    top=1mm,
    bottom=1mm,
    enhanced,
    breakable,
    listing only,
    listing options={
        basicstyle=\ttfamily\small,
        breaklines=true,
        breakatwhitespace=true,
        columns=fullflexible
    }
}
You are an expert knowledge graph curator evaluating the quality of automatically \
generated descriptions for an ontology IRI.

You will be shown:
  1. A database context - a numbered list of RDF triples that describe the IRI and its \
relationships in the knowledge graph. Treat this as a sampled reference, but note that \
it is not necessarily the complete information about the IRI.
  2. {n} candidate descriptions to evaluate, labelled A, B{c_label}.

Rate EACH description independently on all four criteria using a score from 1 to 5. \
Your scores for one description should not be influenced by the scores you give another.

ACCURACY - Does the description correctly represent the concept?
  5 Excellent  All statements are factually correct and expressed with full precision. No misleading, ambiguous, or unsupported claims.
  4 Very Good  All statements are accurate and reasonably precise, with only minor ambiguities that do not mislead.
  3 Good       All statements are plausibly accurate, but at least one is imprecise, overgeneralized, or weakly supported.
  2 Fair       Contains at least one accurate statement, but one or more clearly incorrect claims are also present.
  1 Poor       No accurate statements. All claims are factually incorrect.

CLARITY - Is the description readable and accessible to someone unfamiliar with the ontology?
  5 Excellent  Completely clear, fluent, and unambiguous throughout.
  4 Very Good  Generally clear and easy to read, with only minor phrasing issues that do not impede understanding.
  3 Good       Core meaning is accessible, but at least one phrase is notably ambiguous or awkwardly constructed.
  2 Fair       Difficult to follow; requires significant effort to infer the core meaning.
  1 Poor       Largely incomprehensible; reader cannot extract a coherent meaning.

COMPLETENESS - Does the description include all key information?
  5 Excellent  Covers all essential information and relevant secondary details.
  4 Very Good  Covers all essential information but omits secondary details that would further enrich it.
  3 Good       Covers the core adequately but omits at least one secondary detail expected in an ideal description.
  2 Fair       Covers some essential information but is missing one or more critical details.
  1 Poor       Covers only a superficial aspect; most essential information is absent.

CRUCIALITY - Is all of the included information essential for understanding the concept?
  5 Excellent  Every included detail is essential toward understanding the concept, and nothing is superfluous.
  4 Very Good  Well-focused with at most one minor unnecessary detail.
  3 Good       Mostly focused but includes at least one tangential or redundant detail.
  2 Fair       Right concept is being described, but multiple non-essential details significantly dilute focus.
  1 Poor       Dominated by off-topic content; appears to address a different concept entirely.

---

IRI: {iri}

DATABASE CONTEXT:
{context}

---

{descriptions_block}

---

Respond with a JSON object and nothing else - no explanation, no markdown fences:
{response_schema}
"""
\end{tcblisting}
\captionof{figure}{Prompt Template for Method Comparison (LLM Judge)}
\label{fig:method-comparison-prompt}

\begin{table}[h]
\centering
\caption{Accuracy Scoring Rubric}
\label{tab:rubric-accuracy}
\begin{tabular}{|c|l|p{5cm}|}
\hline
\textbf{Score} & \textbf{Label} & \textbf{Description} \\
\hline
1 & Poor & There are no accurate statements about the term. All the claims that are present in the description are factually incorrect.\\
\hline
2 & Fair & The description contains at least one accurate statement, but one or more clearly incorrect claims are also present.\\
\hline
3 & Good & All statements are plausibly accurate, but at least one is imprecise, overgeneralized, or weakly supported by the available evidence.\\
\hline
4 & Very Good & All statements are accurate and reasonably precise, with only minor ambiguities in phrasing that do not mislead the reader.\\
\hline
5 & Excellent & All statements are factually correct and expressed with full precision. There are no misleading, ambiguous, or unsupported claims.\\
\hline
\end{tabular}
\end{table}

\begin{table}[h]
\centering
\caption{Clarity Scoring Rubric}
\label{tab:rubric-clarity}
\begin{tabular}{|c|l|p{5cm}|}
\hline
\textbf{Score} & \textbf{Label} & \textbf{Description} \\
\hline
1 & Poor & The description is largely incomprehensible and the reader cannot extract a coherent meaning from it.\\
\hline
2 & Fair & The description is difficult enough to follow that it requires significant effort to infer the core meaning. \\
\hline
3 & Good & The core meaning is accessible, but at least one sentence or phrase is notably ambiguous or awkwardly constructed.\\
\hline
4 & Very Good & The description is generally clear and easy to read, with only minor phrasing issues that do not impede understanding.\\
\hline
5 & Excellent & The description is completely clear, fluent, and unambiguous throughout.\\
\hline
\end{tabular}
\end{table}

\begin{table}[h]
\centering
\caption{Completeness Scoring Rubric}
\label{tab:rubric-completeness}
\begin{tabular}{|c|l|p{5cm}|}
\hline
\textbf{Score} & \textbf{Label} & \textbf{Description} \\
\hline
1 & Poor & The description covers only a superficial aspect of the concept and most essential information is absent. \\
\hline
2 & Fair & The description covers some essential information but is missing one or more details that are critical to understanding the concept. \\
\hline
3 & Good & The description covers the core of the concept adequately, but omits at least one secondary detail that would be expected in an ideal description of the term.\\
\hline
4 & Very Good & The description covers all essential information needed to understand the concept, but omits secondary details that would further enrich it.\\
\hline
5 & Excellent & The description covers all essential information and relevant secondary details, giving a full and representative picture of the concept.\\
\hline
\end{tabular}
\end{table}

\begin{table}[h]
\centering
\caption{Cruciality Scoring Rubric}
\label{tab:rubric-cruciality}
\begin{tabular}{|c|l|p{5cm}|}
\hline
\textbf{Score} & \textbf{Label} & \textbf{Description} \\
\hline
1 & Poor & The description is dominated by off-topic or irrelevant content and appears to address a different concept entirely.\\
\hline
2 & Fair & It is clear that the right concept is being described, but multiple non-essential details are present that significantly dilute the focus. \\
\hline
3 & Good & The description is mostly about the right concept, but includes at least one tangential or redundant detail that adds noise without aiding understanding. \\
\hline
4 & Very Good & The description is well-focused with at most one minor unnecessary detail that does not materially harm the overall description. \\
\hline
5 & Excellent & Every included detail is essential toward understanding the concept, and nothing is superfluous.\\
\hline
\end{tabular}
\end{table}

%% file: refs.bib
@misc{w3c-rdfs-comment,
  author       = {Dan Brickley and R. V. Guha},
  title        = {RDF Schema 1.1 — Section 3.7: rdfs:comment},
  year         = {2014},
  howpublished = {\url{https://www.w3.org/TR/rdf-schema/#ch_comment}},
  note         = {W3C Recommendation},
  institution  = {World Wide Web Consortium (W3C)},
}

@misc{owl-spec,
  author       = {Deborah L. McGuinness and Frank van Harmelen},
  title        = {OWL Web Ontology Language Overview},
  year         = {2004},
  howpublished = {\url{https://www.w3.org/TR/owl-features/}},
  note         = {W3C Recommendation, 10 February 2004},
}

@inproceedings{shirvani2025knowledge,
  title={A Knowledge Graph Informing Soil Carbon Modeling},
  author={Shirvani-Mahdavi, Nasim and Wingfield, Devin and Guajardo Gutierrez, Juan and Tran, Mai and Zhu, Zhengyuan and Zhang, Zeyu and Zhang, Haiqi and Goudar, Abhishek Divakar and Li, Chengkai and Jin, Virginia and Propst, Timothy and Roberts, Dan and Stewart, Catherine and Su, Jianzhong and Woodward-Greene, Jennifer},
  booktitle={International Conference on Web Engineering},
  year={2025},
  organization={Springer}
}

@article{dbpedia,
  author    = {Jens Lehmann and Robert Isele and Max Jakob and Anja Jentzsch and Dimitris Kontokostas and Pablo N. Mendes and Sebastian Hellmann and Mohamed Morsey and Patrick van Kleef and Sören Auer and Christian Bizer},
  title     = {DBpedia – A Large-scale, Multilingual Knowledge Base Extracted from Wikipedia},
  journal   = {Semantic Web},
  volume    = {6},
  number    = {2},
  pages     = {167--195},
  year      = {2015},
  publisher = {IOS Press}
}

@article{wikidata,
  author    = {Denny Vrandečić and Markus Krötzsch},
  title     = {Wikidata: A Free Collaborative Knowledge Base},
  journal   = {Communications of the ACM},
  volume    = {57},
  number    = {10},
  pages     = {78--85},
  year      = {2014},
  publisher = {ACM}
}

@article{kaljurand2007verbalizing,
  title={Verbalizing owl in attempto controlled english},
  author={Kaljurand, Kaarel and Fuchs, Norbert E},
  year={2007},
  journal={University of Zurich}
}

@inproceedings{stevens2011automating,
  title={Automating generation of textual class definitions from OWL to English},
  author={Stevens, Robert and Malone, James and Williams, Sandra and Power, Richard and Third, Allan},
  booktitle={Journal of Biomedical Semantics},
  volume={2},
  pages={1--20},
  year={2011},
  organization={Springer}
}

@article{third2011owl,
  title={OWL to English: a tool for generating organised easily-navigated hypertexts from ontologies},
  author={Third, Allan and Williams, Sandra and Power, Richard},
  year={2011}
}

@inproceedings{cregan2007sydney,
  title={Sydney OWL Syntax-towards a Controlled Natural Language Syntax for OWL 1.1.},
  author={Cregan, Anne and Schwitter, Rolf and Meyer, Thomas and others},
  booktitle={OWLED},
  volume={258},
  year={2007}
}

@inproceedings{power2010complexity,
  title={Complexity assumptions in ontology verbalisation},
  author={Power, Richard},
  booktitle={Proceedings of the ACL 2010 Conference Short Papers},
  pages={132--136},
  year={2010}
}

@article{androutsopoulos2013generating,
  title={Generating natural language descriptions from OWL ontologies: the NaturalOWL system},
  author={Androutsopoulos, Ion and Lampouras, Gerasimos and Galanis, Dimitrios},
  journal={Journal of Artificial Intelligence Research},
  volume={48},
  pages={671--715},
  year={2013}
}

@inproceedings{lippolis2025ontology,
  title={Ontology generation using large language models},
  author={Lippolis, Anna Sofia and Saeedizade, Mohammad Javad and Keskis{\"a}rkk{\"a}, Robin and Zuppiroli, Sara and Ceriani, Miguel and Gangemi, Aldo and Blomqvist, Eva and Nuzzolese, Andrea Giovanni},
  booktitle={European Semantic Web Conference},
  pages={321--341},
  year={2025},
  organization={Springer}
}

@article{val2025ontogenix,
  title={OntoGenix: Leveraging Large Language Models for enhanced ontology engineering from datasets},
  author={Val-Calvo, Mikel and Aranguren, Mikel Ega{\~n}a and Mulero-Hern{\'a}ndez, Juan and Almagro-Hern{\'a}ndez, Gin{\'e}s and Deshmukh, Prashant and Bernab{\'e}-D{\'\i}az, Jos{\'e} Antonio and Espinoza-Arias, Paola and S{\'a}nchez-Fern{\'a}ndez, Jos{\'e} Luis and Mueller, Juergen and Fern{\'a}ndez-Breis, Jesualdo Tom{\'a}s},
  journal={Information Processing \& Management},
  volume={62},
  number={3},
  pages={104042},
  year={2025},
  publisher={Elsevier}
}

@article{funk2023towards,
  title={Towards ontology construction with language models},
  author={Funk, Maurice and Hosemann, Simon and Jung, Jean Christoph and Lutz, Carsten},
  journal={arXiv preprint arXiv:2309.09898},
  year={2023}
}

@inproceedings{shi2023hallucination,
  title={Hallucination mitigation in natural language generation from large-scale open-domain knowledge graphs},
  author={Shi, Xiao and Zhu, Zhengyuan and Zhang, Zeyu and Li, Chengkai},
  booktitle={Proceedings of the 2023 Conference on Empirical Methods in Natural Language Processing},
  pages={12506--12521},
  year={2023}
}

@article{zhang2023enhancing,
  title={Enhancing RDF Verbalization with Descriptive and Relational Knowledge},
  author={Zhang, Fan and Zhang, Meishan and Liu, Shuang and Sun, Yueheng and Duan, Nan},
  journal={ACM Transactions on Asian and Low-Resource Language Information Processing},
  volume={22},
  number={6},
  pages={1--18},
  year={2023},
  publisher={ACM New York, NY}
}

@inproceedings{du2024bi,
  title={Bi-Directional Multi-Granularity Generation Framework for Knowledge Graph-to-Text with Large Language Model},
  author={Du, Haowei and Li, Chen and Zhang, Dinghao and Zhao, Dongyan},
  booktitle={Proceedings of the 62nd Annual Meeting of the Association for Computational Linguistics (Volume 2: Short Papers)},
  pages={147--152},
  year={2024}
}

@misc{nsf_proto_okn,
  title        = {NSF 23-571: Building the Prototype Open Knowledge Network (Proto-OKN)},
  author       = {{National Science Foundation}},
  howpublished = {\url{https://www.nsf.gov/funding/opportunities/proto-okn-building-prototype-open-knowledge-network/506169/nsf23-571/solicitation}},
  year         = {2023},
  note         = {Accessed: 2026-03-03}
}

@misc{biobricks_mesh,
  author       = {{RENCI Open Knowledge Network}},
  title        = {BioBricks {MeSH} Knowledge Graph},
  year         = {2023},
  howpublished = {\url{https://frink.renci.org/registry/kgs/biobricks-mesh/}},
  note         = {Accessed: 2026-04-29}
}

@article{10.5555/2786071.2786081,
author = {Isaac, Antoine and Haslhofer, Bernhard},
title = {Europeana Linked Open Data --data.europeana.eu},
year = {2013},
issue_date = {July 2013},
publisher = {IOS Press},
address = {NLD},
volume = {4},
number = {3},
issn = {1570-0844},
journal = {Semant. Web},
month = jul,
pages = {291–297},
numpages = {7}
}

@article{10.1093/database/baab069,
    author = {Jackson, Rebecca and Matentzoglu, Nicolas and Overton, James A and Vita, Randi and Balhoff, James P and Buttigieg, Pier Luigi and Carbon, Seth and Courtot, Melanie and Diehl, Alexander D and Dooley, Damion M and Duncan, William D and Harris, Nomi L and Haendel, Melissa A and Lewis, Suzanna E and Natale, Darren A and Osumi-Sutherland, David and Ruttenberg, Alan and Schriml, Lynn M and Smith, Barry and Stoeckert Jr., Christian J and Vasilevsky, Nicole A and Walls, Ramona L and Zheng, Jie and Mungall, Christopher J and Peters, Bjoern},
    title = {OBO Foundry in 2021: operationalizing open data principles to evaluate ontologies},
    journal = {Database},
    volume = {2021},
    pages = {baab069},
    year = {2021},
    month = {09},
    issn = {1758-0463},
    doi = {10.1093/database/baab069},
    url = {https://doi.org/10.1093/database/baab069},
    eprint = {https://academic.oup.com/database/article-pdf/doi/10.1093/database/baab069/40854912/baab069.pdf},
}

@article{Zaitoun_Sagi_Peleg_2024, title={Generating Ontology-Learning Training-Data through Verbalization}, volume={4}, url={https://ojs.aaai.org/index.php/AAAI-SS/article/view/31797}, DOI={10.1609/aaaiss.v4i1.31797}, abstractNote={Ontologies play an important role in the organization and representation of knowledge. However, in most cases, ontologies do not fully cover domain knowledge, resulting in a gap. This gap, often expressed as a lack of concepts, relations, or axioms, is usually filled by domain experts in a manual and tedious process. Utilizing large language models (LLMs) can ease this process; a fine-tuned LLM could receive as input up-to-date and reliable domain knowledge natural text and output a structured graph in OWL RDF/Turtle format, which is the standard format of ontologies. Thus, to fine-tune a model, text-owl sentence pairs that constitute such a dataset must be acquired. Unfortunately, such a dataset does not exist in the literature or within the open-source community. Therefore, this paper introduces our LLM-assisted verbalizer to create such a data set by converting OWL statements from existing ontologies into natural text. We evaluate the verbalizer on 322 classes from four different ontologies using two different LLMs, achieving precision and recall as high as 0.99 and 0.96, respectively.}, number={1}, journal={Proceedings of the AAAI Symposium Series}, author={Zaitoun, Antonio and Sagi, Tomer and Peleg, Mor}, year={2024}, month={Nov.}, pages={233–241} }

@article{kuhn-2014-survey,
    title = "A Survey and Classification of Controlled Natural Languages",
    author = "Kuhn, Tobias",
    journal = "Computational Linguistics",
    volume = "40",
    number = "1",
    month = mar,
    year = "2014",
    address = "Cambridge, MA",
    publisher = "MIT Press",
    url = "https://aclanthology.org/J14-1005/",
    doi = "10.1162/COLI_a_00168",
    pages = "121--170"
}

@misc{openai2025gpt41mini,
  author       = {{OpenAI}},
  title        = {GPT-4.1 Mini Model Documentation},
  year         = {2025},
  howpublished = {\url{https://platform.openai.com/docs/models/gpt-4.1-mini}},
  note         = {Accessed: 2026-05-08}
}

@misc{google2025gemini25flashlite,
  author       = {{Google DeepMind}},
  title        = {Gemini 2.5 Flash-Lite Model Card},
  year         = {2025},
  howpublished = {\url{https://ai.google.dev/gemini-api/docs/models/gemini-2.5-flash-lite}},
  note         = {Accessed: 2026-05-08}
}

@misc{zhipu2026glm47flash,
  author       = {{Zhipu AI and THUDM}},
  title        = {GLM-4.7 Flash Model Documentation},
  year         = {2026},
  howpublished = {\url{https://github.com/THUDM/GLM}},
  note         = {Accessed: 2026-05-08}
}

@article{bai2026kimik25,
  author  = {Bai, Tongtong and others},
  title   = {Kimi K2.5: Visual Agentic Intelligence},
  journal = {arXiv preprint arXiv:2602.02276},
  year    = {2026},
  doi     = {10.48550/arXiv.2602.02276}
}

@misc{cohere_rerank_3_5_2024,
  title        = {Introducing Rerank 3.5: Precise AI Search},
  author       = {{Cohere}},
  year         = {2024},
  url          = {https://cohere.com/blog/rerank-3pt5},
  note         = {Accessed: 2026-05-08}
}

@inproceedings{wei2022chain,
  title={Chain-of-thought prompting elicits reasoning in large language models},
  author={Wei, Jason and Wang, Xuezhi and Schuurmans, Dale and Bosma, Maarten and Xia, Fei and Chi, Ed and Le, Quoc V and Zhou, Denny and others},
  booktitle={Advances in Neural Information Processing Systems},
  volume={35},
  pages={24824--24837},
  year={2022}
}
